\documentclass[runningheads]{llncs}
\usepackage[T1]{fontenc}
\usepackage{amssymb}
\usepackage{amsmath}
\usepackage[linesnumbered,ruled,vlined]{algorithm2e}
\usepackage{booktabs}
\usepackage{graphicx}
\usepackage{textcomp}
\usepackage{xcolor}
\usepackage{color}
\usepackage[binary-units]{siunitx}
\usepackage{multicol}
\usepackage{multirow}
\usepackage{bm}
\usepackage{subcaption}
\usepackage{url}
\usepackage{autobreak}
\usepackage[para]{footmisc}
\usepackage{makecell}
\usepackage{xspace}
\usepackage{comment}
\usepackage{threeparttable}
\usepackage{eso-pic}

\AddToShipoutPictureFG*{%
\AtPageUpperLeft{%
\raisebox{-1.5cm}[0pt][0pt]{%
\makebox[\paperwidth]{%
\hfill
\textcolor{red}{This is the first submitted version of the manuscript.}
\hfill
}}}}
\newcommand{\Malaga}{M\'{a}laga\xspace}
\begin{document}
\title{A Methodology for Effective Surrogate Learning in Complex Optimization}
\titlerunning{A Methodology for Effective Surrogate Learning in Complex Optimization}
%
\author{Tomohiro Harada\inst{1}\orcidID{0000-0002-0704-4351} \and
Enrique Alba\inst{2}\orcidID{0000-0002-5520-8875} \and Gabriel Luque\inst{2}\orcidID{0000-0001-7909-1416}    }
%
\authorrunning{T. Harada, E. Alba, and G. Luque}
%
\institute{Saitama University, 255 Shimo-okubo, Sakura-ku, Saitama City, 3388570 Saitama, Japan \email{tharada@mail.saitama-u.ac.jp} \and
ITIS University of Malaga, 29071 Malaga, Spain
\email{ealbat@uma.es}, \email{gluque@uma.es}}
%
\maketitle              
\begin{abstract}
Solving complex problems requires continuous effort in developing theory and practice to cope with larger, more difficult scenarios. Working with surrogates is normal for creating a proxy that realistically models the problem into the computer. Thus, the question of how to best define and characterize such a surrogate model is of the utmost importance. In this paper, we introduce the PTME methodology to study deep learning surrogates by analyzing their Precision, Time, Memory, and Energy consumption. We argue that only a combination of numerical and physical performance can lead to a surrogate that is both a trusted scientific substitute for the real problem and an efficient experimental artifact for scalable studies. Here, we propose different surrogates for a real problem in optimally organizing the network of traffic lights in European cities and perform a PTME study on the surrogates' sampling methods, dataset sizes, and resource consumption. We further use the built surrogates in new optimization metaheuristics for decision-making in real cities. We offer better techniques and conclude that the PTME methodology can be used as a guideline for other applications and solvers.

\end{abstract}
\section{Introduction}

Surrogate models are increasingly employed to accelerate optimization and learning in domains where evaluating candidate solutions is costly~\cite{BLIEK2023110744,HWANG201874,WASEEM2025344}. Many real-world problems involve simulation-based evaluations, non-linear domains, uncertainty, and complex constraints in mixed discrete and continuous spaces. The resulting intricate internal dynamics, combined with typical high dimensionality, push researchers to explore better model designs and evaluation protocols when developing algorithms for real problems.

When it comes to non-deterministic solvers, just the step of obtaining one fitness value for a tentative solution may require running a detailed digital twin, often including some kind of microscopic simulator~\cite{Balderas2021,WASEEM2025344}, that can represent the problem accurately enough for the results to be used in practice. Surrogates, typically based on machine learning, such as deep neural networks (DNNs), approximate the input-output behavior of these costly complex systems. Once trained, they allow optimization algorithms to explore the search space at a controlled cost~\cite{cheng2024review}. This domain is seeking ways to achieve better cost efficiency and numerical accuracy in the solve, such as it happens, e.g., for metaheuristics~\cite{Harada2024Energy}, showing that the phases of training and inference can be decoupled and managed in new ways, thus paving the way for modern techniques. 

However positive all this is, the adoption of a surrogate model introduces new methodological challenges that have been hardly addressed in the literature when considering the full picture. Most existing works evaluate surrogates in terms of the model error~\cite{DAVIS2018979,Harada2023} or running times~\cite{BLIEK2023110744,HWANG201874}, thus overlooking other critical dimensions such as the memory footprint or the energy consumption. This could lead to incomplete and sometimes misleading conclusions. In today's research, not only numerical precision matters, but physical performance (time and memory)~\cite{khalfi2023metaheuristics} and sustainability (notably, energy) matter~\cite{Abdelhafez2019,Jamil2022,Paul2023Green}. 

To address this gap, we introduce PTME, a methodological evaluation protocol for analysing surrogate models along four complementary axes: Precision, Time, Memory, and Energy. PTME provides a unified evaluation framework to ease the design, comparison, and real use of surrogates. We will illustrate here this approach by considering both numerical performance (prediction accuracy and ranking reliability) and computational performance (resource usage during training and inference). The integration of results is here natural, easy for other researchers, and more comprehensive to really answer the question of what is a good surrogate for a real application.

Our use case will be a real-world traffic scheduling optimization task, where evaluating candidate city plans requires high-fidelity microscopic simulation (vehicles, pedestrians, driving rules, real city data...). Our results reveal meaningful tradeoffs between accuracy and computational sustainability, and illustrate how PTME can help.

The remainder of the paper is organised as follows. Section~\ref{sec:PTME} presents the PTME methodology. Section~\ref{sec:case_study} describes the case study, outlining the problem and the surrogate design. Section~\ref{sec:experiment} details the experimental setting, while Section~\ref{sec:surrogate} analyses the behavior of the surrogate model under PTME. Section~\ref{sec:search} focuses on the surrogate-assisted solvers, and Section~\ref{sec:conclusion} summarises the work.

\section{PTME Methodology}\label{sec:PTME}
This section introduces the PTME methodology, a framework for evaluating surrogate models along four complementary dimensions: Precision, Time, Memory, and Energy. In fact, PTME is a concrete way in which we can balance numerical and physical performances, two of the most important factors determining the suitability of a surrogate (others are possible).

\subsection{PTME Dimensions and Measurements}\label{sec:PTME_dim}

Let us here discuss the two main dimensions first, and later how we could quantify them. As to precision, it refers to how accurately the surrogate approximates the true objective function generated by a complex problem (often obtained with costly simulations), as inaccurate models may mislead a further optimization phase. Precision will be assessed here by using error (the common loss function) and ranking metrics, the latter capturing consistency in sorting tentative solutions according to their suitability for the problem, which is essential, e.g., in swarm algorithms.
We analyse our use case with the mean absolute percentage error (MAPE), the root mean squared error (RMSE), and Kendall's $\tau$ ranking. As to the error metrics MAPE and RMSE, they are formulated as follows:

\begin{minipage}[b]{0.48\linewidth}
\begin{equation}
\mathrm{MAPE} = \frac{1}{n} \sum_{i=1}^n \left| \frac{\hat{y}_i - y_i}{y_i} \right| \times 100,
\end{equation}
\end{minipage}
\hfill
\begin{minipage}[b]{0.48\linewidth}
\begin{equation}
\mathrm{RMSE} = \sqrt{\frac{1}{n} \sum_{i=1}^n \left( \hat{y}_i - y_i \right)^2 },
\end{equation}
\end{minipage}
where $y_i$ and $\hat{y}_i$ represent the true and predicted objective values for $n$ observations. Smaller MAPE and RMSE values indicate higher prediction, with a value of $0$ indicating perfect matching.
MAPE expresses errors as relative percentages, making it intuitive and scale-independent but unstable when true values approach zero, whereas RMSE penalizes large deviations more strongly, providing sensitivity to outliers but depending on the data’s absolute scale.

Kendall's $\tau$ comes in two main forms: $\tau_a$, which does not adjust for ties, while $\tau_b$ does so. They are defined as follows:\\
\begin{minipage}[b]{0.3\linewidth}
\begin{equation}
    \tau_a=\frac{N_c-N_d}{\tfrac{1}{2}n(n-1)},
\end{equation}
\end{minipage}
\hfill
\begin{minipage}[b]{0.68\linewidth}
\begin{equation}
    \tau_b = \frac{N_c - N_d}{\sqrt{(N_c+N_d+T_x)(N_c+N_d+T_y)}},
\end{equation}
\end{minipage}
where $N_c$ and $N_d$ represent the numbers of concordant and discordant pairs. $T_x$ and $T_y$ denote the total number of ties in the predicted or true rankings, respectively, and $\tau_b$ adjusts the denominator to account for ties. 
A value close to $+1$ indicates strong agreement in rankings, a value around $0$ suggests no correlation, and a value close to $-1$ implies complete disagreement. 

Time captures the computational effort for model training and inference. Training time determines the cost of constructing or updating the surrogate, while inference time limits how many candidate solutions can be evaluated during optimization.

Memory refers to the storage and runtime footprint of the surrogate model, which is particularly relevant for deep learning. Memory usage may be characterized by model size, peak allocation during execution, and activation footprint during inference.

Energy quantifies the power consumption of training and inference. As computation costs and sustainability concerns increase, energy usage becomes an important dimension. It can be estimated using hardware power counters, system-level monitors, or software-based energy measurement tools, and may be reported directly or converted into carbon impact estimates.

In short, we stress the importance of including all these dimensions in every single study, since not doing so makes a weaker contribution to surrogate studies, and then their help in real decision making system will be delayed.

\subsection{Measurement Protocol}

The PTME methodology requires evaluating each dimension under controlled and comparable conditions. Since training and inference have different computational behavior and implications for the later surrogate-based optimization, both phases should be assessed independently as a first step. In general, the protocol involves: (i) defining how the surrogate will be trained, (ii) specifying how inference will be evaluated, and (iii) establishing a consistent procedure for collecting and aggregating the corresponding measurements.

For precision, evaluation metrics are obtained by comparing surrogate outputs with those of the original problem over a set of representative samples of the search space. For time, memory, and energy measurements, they should be collected during both training and inference in a reproducible manner. These metrics can be taken through system-level monitoring tools, hardware counters, or software instrumentation, depending on the computational environment.


\begin{algorithm}[htb]
\caption{PTME measuring procedure for training and inference}
\label{alg:measure_ann}
\DontPrintSemicolon
$E_{CPU,t}, E_{DRAM,t}, T_t, M_t \gets \emptyset$ \tcp*{Subscript $t$ indicates training}
$E_{CPU,i}, E_{DRAM,i}, T_i, M_i \gets \emptyset$ \tcp*{Subscript $i$ indicates inference}
$R \gets \emptyset$ \tcp*{Stores pairs of true and predicted fitness values}
Sample $N_{test}$ test data and evaluate their true objective values\tcp*{Common test data}
\For{$i_t = 1$ \KwTo $T$}{
    Sample $N_{train}$ training data and evaluate their true objective values\;
    Train DNN with the dataset while measuring computational metrics\;
    Record computational metrics into $E_{CPU,t}$, $E_{DRAM,t}$, $T_t$ and $M_t$\;
    \For{$i_i = 1$ \KwTo $N_{test}$}{
        Predict the objective value of the $i_i$-th test sample while measuring computational metrics\;
        Record computational metrics $E_{CPU,i}$, $E_{DRAM,i}$, $T_i$, and $M_i$\;
        Record true and predicted objective value pair into $R$\;
    }
}
Compute statistics of all computational metrics\;
Compute $MAPE$, $RMSE$, and Kendall's $\tau$ using $R$\;
\end{algorithm}
Algorithm~\ref{alg:measure_ann} outlines an example procedure for measuring PTME of surrogate models during training and inference phases. This procedure begins by generating a test dataset of size $N_{test}$ while evaluating their true objective values. It then repeats the sampling, training, and inference steps $T$ times, generating a new training dataset of size $N_{train}$ each time. The training and test data are sampled using stochastic or deterministic methods, such as random sampling.

Then, the surrogate is trained with the sampled dataset while measuring CPU and DRAM energy, time, and memory usage. The measured values are stored in the corresponding lists for later analysis.
After training, the surrogate is evaluated on a common test dataset of size $N_{test}$. The test dataset is held constant across evaluations to ensure consistency. For each test sample, the surrogate predicts the objective value while measuring the same computational metrics as those used during training. The true and predicted objective values are recorded for later calculation of precision metrics.
After completing all repetitions, the statistics of PTME are computed using the recorded measurements.



\section{Case Study: Traffic-Light Optimization in Real Cities}\label{sec:case_study}
This section describes the case study used to demonstrate the PTME methodology. We first outline the real-world optimization problem (traffic-light scheduling problem) and then face two important steps in designing a surrogate for it: (i) the study to sampling strategies used to generate the training dataset to design surrogate models (since the search space is immense) and (ii) the DNN architecture.
\subsection{Traffic-Light Scheduling: A Costly Real-World Problem}\label{sec:problem}
This study focuses on algorithms and surrogate modeling, motivated by the need to solve computationally expensive problems. We then selected a challenging yet practical real-world task: optimizing traffic lights for sustainable urban mobility. Traffic lights indeed play a crucial role in managing traffic flow at intersections by cycling through signal phases. Each phase includes a combination of coloured lights and allows vehicles to use the roadway for a certain duration. A complete assignment of phase durations across all intersections defines a traffic-light plan, and optimizing it is critical to reducing stops, delays, and travel time.

The optimization problem involves multiple objectives, and to represent this problem mathematically, we adopt the formulation of \cite{GARCIANIETO2012274}. The goal is to maximize the number of vehicles arriving at their destination ($NV_D$) and minimize the number of vehicles that do not reach their destination ($NV_{ND}$) during the simulation time ($T_S$). We also minimize the total travel time of all vehicles ($TT_v$) and their total waiting time ($TT_{EP}$). Furthermore, we maximize the ratio of green to red durations across all intersections, defined as
$P =\sum_{i=0}^{inter}\sum_{j=0}^{fs}d_{i,j}g_{i,j}/r_{i,j}$, where $inter$ and $fs$ represent the number of intersections and phases in each intersection, $g_{i,j}$ and $r_{i,j}$ denote the number of green and red signals at intersection $i$ and phase state $j$, and $d_{i,j}$ represents the duration of these signals. 

These objectives are combined into a single objective function as:
\begin{equation}
F= \frac{TT_{v} + TT_{EP} + NV_{ND} \times T_{S}}{NV_{D}^{2}+P}.\label{eq:prob2}
\end{equation}

The goal is to minimize Eq.~\eqref{eq:prob2}. A candidate solution is a positive integer vector specifying the duration of each traffic light phase at each intersection in seconds. 

To evaluate a candidate solution, we need to simulate the corresponding traffic dynamics using SUMO~\cite{SUMO2018}. While the results of the simulator are highly accurate and closely reflect reality, the process is time-consuming, taking several seconds to tens of minutes depending on the city size. This makes the problem computationally expensive, motivating the need for the use of surrogate models.

The experimental evaluation considers three real-world traffic light scheduling instances from \Malaga, Stockholm, and Paris, each characterized by different network sizes and traffic conditions shown in Table~\ref{tab:problem_instances}.
These settings are based on several previous studies~\cite{Segredo2019,VILLAGRA2020101085} that addressed the same problem instances. 

\begin{table}[tb]
\centering
\caption{Information about the problem instances}
\label{tab:problem_instances} 
\scriptsize{
\begin{tabular}{lrrr}
\toprule
     &  \Malaga  & Stockholm &Paris\\
 \midrule
Total number of intersections& 56 &75&70\\
Total number of phases& 190 &370&378\\
Total number of vehicles&   1,200&  1,400&1,200\\
Simulation time (\si{\second})& 2,200   &4,000&3,400\\
\bottomrule
\end{tabular}
}
\vspace{-2em}
\end{table}

\subsection{Sampling Strategies for Surrogate Training}
The performance of surrogate models strongly depends on the quality of the training data. Since only a limited number of evaluations can be performed, sampling strategies that cover the search space effectively are crucial. To investigate their impact on surrogates, we consider two widely used sampling methods: uniform random sampling (URS) and Latin hypercube sampling (LHS).

URS draws the $j$-th decision variable of the $i$-th sample, $x_j^i$, independently from a uniform distribution over the feasible domain $\Omega = [a, b]^D\subset \mathbb{R}^D$, where $D$ denotes the problem dimensionality, and $[a, b]$ represents the lower and upper bounds for each variable. While URS is simple and unbiased, it may produce clustered samples or leave regions unexplored, especially when the sample size is small relative to problem dimensionality.

LHS~\cite{Fang2005} improves coverage by stratifying each dimension into $N$ equal probability intervals and selecting one sample uniformly within each interval. Formally, for $N$ samples and dimension $j$, let $\pi_j$ be a random permutation of $\{1,\dots,N\}$; then
\begin{equation}
    x_j^i=a+(b-a)\cdot\frac{\pi_j(i)-u_j^i}{N}, \quad u_j^i\sim U(0,1).
\end{equation}
This stratified design yields better space-filling properties than URS.

By comparing URS with LHS, we aim to evaluate how sample distribution affects the accuracy of surrogates and the computational efficiency of surrogate-assisted optimization.
Following prior works~\cite{Segredo2019,VILLAGRA2020101085}, we set $a=4$ and $b=60$ for odd-indexed variables and fix even-indexed variables at $4$. 

\subsection{Deep Neural Network-Based Surrogate Architecture}

\begin{table}[tb]
\begin{minipage}[t]{0.48\linewidth}
\centering
\caption{DNN surrogate architecture}
\label{tb:ann_arch}
\scriptsize{
\begin{tabular}{lccc}
\toprule
&\multicolumn{3}{c}{Value}\\
\cmidrule(lr){2-4}
Parameter&\Malaga&Stockholm&Paris\\
\midrule
\# inputs&190&370&378\\
\# hidden layers&2&2&2\\
\# hidden neurons&285--190&555--370&567--378\\
Activation function&\multicolumn{3}{c}{ReLU}\\
\bottomrule
\end{tabular}
}
\end{minipage}
\hfill
\begin{minipage}[t]{0.48\linewidth}
\centering
\caption{Training parameters}
\label{tb:param_ann}
\scriptsize{
\begin{tabular}{lc}
\toprule
Parameter&Value\\
\midrule
Optimizer&Adam\\ 
\# training epochs& 100\\
Batch size&32\\
Learning rate&$1.0\times 10^{-4}$\\
Loss function&Mean squared error (MSE)\\
\bottomrule
\end{tabular}
}
\end{minipage}
\end{table}
This study employs DNNs as surrogate models due to their ability to capture complex, non-linear relationships in high-dimensional data.
Table~\ref{tb:ann_arch} summarizes the DNN architecture used for the surrogates, adopting the same architectural setting as described in~\cite{Harada2024Energy}. The number of inputs corresponds to the problem dimensionality: 190 for \Malaga, 370 for Stockholm, and 378 for Paris. 
The DNN consists of two hidden layers. The first hidden layer contains 1.5 times the number of input neurons, and the second layer matches the number of input neurons. For instance, the \Malaga instance has hidden layers of 285 and 190 neurons, respectively. Similarly, Stockholm and Paris have hidden layers of 555--370 and 567--378 neurons, respectively.
The hidden layers use the ReLU activation function. 
Table~\ref{tb:param_ann} shows their training parameters. DNN is trained with the Adam optimizer for 100 epochs, using a batch size of 32 and a learning rate of $1.0\times 10^{-4}$. The training loss function is mean squared error (MSE).

\section{Experimental Setup and Metrics}\label{sec:experiment}

All experiments were conducted on a computer running Ubuntu 22.04 with an Intel(R) Xeon(R) CPU E5-1650 v2 operating at \SI{3.50}{\giga\hertz} and \SI{16}{\giga\byte} of DRAM.
The algorithms were implemented in Python 3.10.12. DNN surrogates were developed using Tensorflow 2.17.0\footnote{\url{https://www.tensorflow.org/}} and Keras 3.5\footnote{\url{https://keras.io/}}. 


To measure energy and time, we used the Python package pyRAPL\footnote{\url{https://pyrapl.readthedocs.io/en/latest/}}, which leverages Intel's Running Average Power Limit (RAPL) technology~\cite{David2010}. Intel RAPL is a reliable standard for energy analysis~\cite{Abdelhafez2019,anthony2020carbontracker,Diaz2022,Ferro2023}. 

For memory usage analysis, we used Python's built-in \texttt{tracemalloc} module. In addition, since \texttt{tracemalloc} does not capture the memory usage of subprocesses, the memory usage of SUMO is measured using the \texttt{time -v} command.

Precision quality of surrogate models is evaluated using MAPE, RMSE, and two types of Kendall's $\tau$ described in Section~\ref{sec:PTME_dim}.
In the experimental procedure shown in Algorithm~\ref{alg:measure_ann}, we set $T=10$ and $N_{test}=10,000$. The test data in line 4 is selected using URS, while the training data in line 6 is selected with either URS or LHS. To evaluate the effect of sampling on model performance and resource consumption, the dataset size $N_{train}$ is varied among 1,000 (1k), 10,000 (10k), 100,000 (100k), and 1,000,000 (1M).

\section{Surrogate Design: Training and Inference Analysis}\label{sec:surrogate}
This section presents the experimental results of training and inference of DNN surrogate models using different sampling strategies and dataset sizes. First, the characteristics of the training dataset are analysed, and then analyses of PTME for both training and inference are provided.

\subsection{Training Data Distribution Analysis}
\begin{figure}[tb]
    \begin{minipage}[c]{0.30\linewidth}
    \centering
        \includegraphics[width=\linewidth]{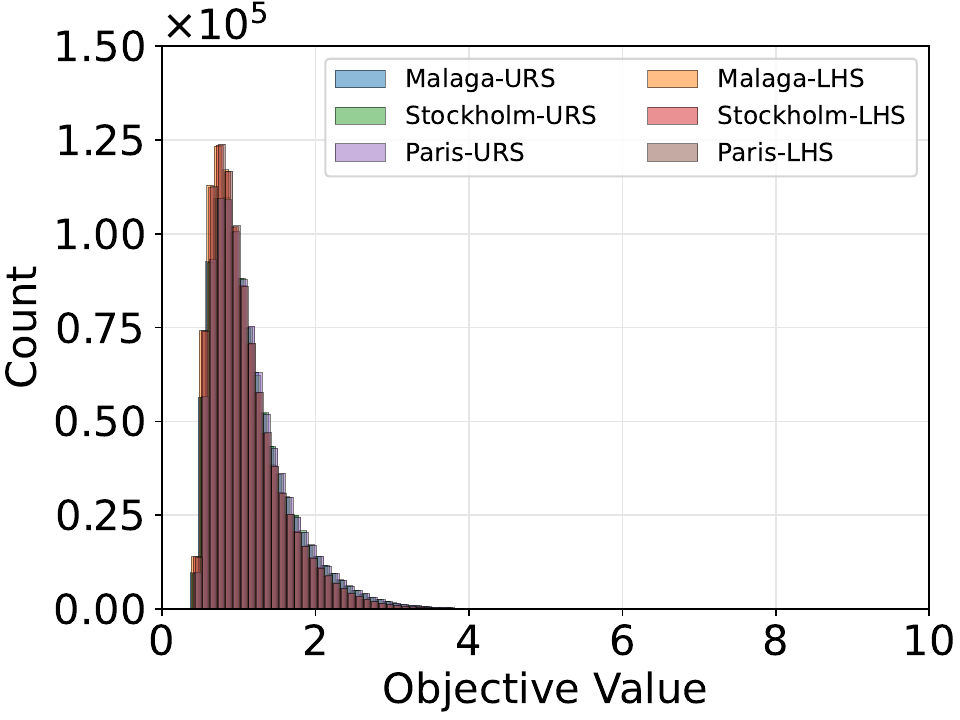}
    \caption{Objective value distribution of 1M dataset}
    \label{fig:fitness_distribution}
    \end{minipage}
    \hfill
    \begin{minipage}[c]{0.68\linewidth}
\centering
\captionof{table}{Mean and variance of objective distribution obtained with the log-normal fitting}
\label{tab:results_mean_var}
\begin{threeparttable}
\fontsize{6}{6}\selectfont{
\begin{tabular}{llrr@{\quad}rr@{\quad}rr@{\quad}rr}
\toprule
\multirow{2}{*}{Instance}&\multirow{2}{*}{Method} & \multicolumn{2}{c}{1k} & \multicolumn{2}{c}{10k} & \multicolumn{2}{c}{100k} & \multicolumn{2}{c}{1M} \\
 &&   Mean & Var. & Mean & Var. & Mean & Var. & Mean & Var. \\
\midrule
\multirow{2}{*}{\Malaga} & URS & 1.15 & 0.24 & 1.15 & 0.23 & 1.15 & 0.23 & 1.15 & 0.23 \\
 & LHS & \textbf{1.08} & \textbf{0.19} & \textbf{1.07} & \textbf{0.19} & \textbf{1.07} & \textbf{0.19} & \textbf{1.07} & \textbf{0.19} \\
\midrule
\multirow{2}{*}{Stockholm} & URS & 1.16 & 0.24 & 1.15 & 0.23 & 1.15 & 0.23 & 1.15 & 0.23 \\
 & LHS & \textbf{1.07} & \textbf{0.19} & \textbf{1.07} & \textbf{0.19} & \textbf{1.07} & \textbf{0.19} & \textbf{1.07} & \textbf{0.19} \\
\midrule
\multirow{2}{*}{Paris} & URS & 1.15 & 0.23 & 1.15 & 0.23 & 1.15 & 0.23 & 1.15 & 0.23 \\
 & LHS & \textbf{1.07} & \textbf{0.19} & \textbf{1.07} & \textbf{0.19} & \textbf{1.07} & \textbf{0.19} & \textbf{1.07} & \textbf{0.19} \\
\bottomrule
\end{tabular}
\begin{tablenotes}
\item[*] The significantly small values between the two sampling methods are highlighted in \textbf{bold}.
\end{tablenotes}
}
\end{threeparttable}
\end{minipage}
\end{figure}
This section provides an overview of the data distribution of the training datasets used in the experiments. 

Fig.~\ref{fig:fitness_distribution} illustrates the distribution of objective values for the 1M dataset. The horizontal axis represents objective values, and the vertical axis denotes their frequency of occurrence. Overall, the distributions appear similar across different dataset sizes and sampling methods, indicating no significant shift in the overall fitness landscape. Consistent patterns are also observed for the other dataset sizes, confirming that this behavior is not specific to the 1M case.

Table~\ref{tab:results_mean_var} reports the arithmetic mean and variance of the objective values obtained from fitting a log-normal distribution. From this analysis, the mean value for URS is approximately 1.15 with a variance of 0.23--0.24, whereas LHS yields a lower mean of around 1.07--1.08 with a variance of 0.19. This indicates that LHS tends to sample higher-quality solutions compared to URS.

Next, we analyse the uniformity of the decision variables in the datasets. To quantify this, we compute the average entropy of each variable for each dataset. If the distribution were perfectly uniform over the defined domain $[4,60]^D$, the theoretical maximum average entropy would be 5.83.
The entropy values remain consistent across all instances. For URS, the entropy is 5.74 for the 1k dataset and increases slightly to 5.78 for larger datasets. In contrast, LHS maintains an entropy of 5.78 across all dataset sizes. Although the numerical difference is small, LHS consistently achieves higher entropy than URS, indicating more uniform coverage of the decision space.

Across all dataset sizes, URS and LHS exhibit similar objective-value distributions. Although LHS produces slightly lower objective values, the difference remains minimal. Regarding decision-space coverage, LHS shows marginally higher entropy for small datasets, whereas the discrepancy becomes negligible as the dataset size increases. Overall, both sampling strategies generate comparable datasets in large-scale settings, with only minor differences in uniformity and objective-value statistics.

\subsection{Training PTME Analysis}
\begin{table}[!tb]
\centering
\caption{Computational metrics of DNN training for different dataset sizes}
\label{tab:results_training}
\begin{threeparttable}
\fontsize{6}{6}\selectfont{
\begin{tabular}{lllrr@{\quad}rr@{\quad}rr@{\quad}rr}
\toprule
\multirow{2}{*}{Instance}&\multirow{2}{*}{Method} & \multirow{2}{*}{Metric} & \multicolumn{2}{c}{1k} & \multicolumn{2}{c}{10k} & \multicolumn{2}{c}{100k} & \multicolumn{2}{c}{1M} \\
 &&&   Avg. & Stdv. & Avg. & Stdv. & Avg. & Stdv. & Avg. & Stdv. \\
\midrule
\multirow{8}{*}{\Malaga}&\multirow{4}{*}{URS} & CPU (\si{\joule}) & \underline{549.51} & 3.23 & 3276.53 & 19.93 & 30691.07 & 215.58 & 291625.29 & 16705.74 \\
& & DRAM (\si{\joule}) & \underline{29.80} & 2.35 & 282.47 & 8.44 & 3185.50 & 61.53 & 29028.86 & 4395.30 \\
& & Time (\si{\second}) & \underline{11.63} & 0.13 & 121.99 & 3.89 & 1394.86 & 14.71 & 11242.17 & 3224.97 \\
& & Memory (\si{\mega\byte}) & \underline{8.86} & 0.27 & 45.29 & 1.61 & \textbf{437.13} & 0.31 & 4364.33 & 0.34 \\
\cmidrule{2-11}
&\multirow{4}{*}{LHS} & CPU (\si{\joule}) & \underline{551.31} & 3.04 & 3253.01 & 38.01 & \textbf{30458.28} & 161.12 & 295883.77 & 18344.62 \\
& & DRAM (\si{\joule}) & \underline{30.06} & 1.20 & 276.27 & 8.14 & 3153.00 & 39.57 & 30760.71 & 3341.39 \\
& & Time (\si{\second}) & \underline{11.62} & 0.06 & 122.02 & 3.61 & 1389.43 & 19.86 & 11349.27 & 3270.30 \\
& & Memory (\si{\mega\byte}) & \underline{8.69} & 0.05 & 45.37 & 1.67 & 437.19 & 0.23 & 4364.14 & 0.01 \\
\midrule
\multirow{8}{*}{Stockholm}&\multirow{4}{*}{URS} & CPU (\si{\joule}) & \underline{856.30} & 12.58 & 6103.56 & 62.90 & 59566.85 & 530.72 & 601370.76 & 17159.39 \\
& & DRAM (\si{\joule}) & \underline{65.00} & 0.98 & 497.29 & 9.28 & 5220.26 & 117.69 & 49783.25 & 7551.17 \\
& & Time (\si{\second}) & \underline{15.50} & 0.08 & 102.64 & 1.20 & 1065.45 & 8.97 & 10559.67 & 395.28 \\
& & Memory (\si{\mega\byte}) & \underline{12.46} & 0.60 & 86.62 & 1.77 & 848.51 & 0.00 & 8484.22 & 0.36 \\
\cmidrule{2-11}
&\multirow{4}{*}{LHS} & CPU (\si{\joule}) & \underline{855.66} & 10.95 & \textbf{6038.10} & 50.02 & 59291.05 & 484.66 & 600296.01 & 13953.13 \\
& & DRAM (\si{\joule}) & \underline{65.41} & 2.17 & 490.42 & 7.74 & 5169.73 & 83.79 & 49016.19 & 10244.93 \\
& & Time (\si{\second}) & \underline{15.50} & 0.09 & \textbf{101.53} & 0.75 & 1061.14 & 9.02 & 10457.43 & 559.92 \\
& & Memory (\si{\mega\byte}) & \underline{12.52} & 0.50 & 86.48 & 1.61 & 848.51 & 0.00 & 8484.16 & 0.33 \\
\midrule
\multirow{8}{*}{Paris}&\multirow{4}{*}{URS} & CPU (\si{\joule}) & \underline{863.34} & 11.64 & 6253.91 & 30.81 & 60924.71 & 327.00 & 622654.85 & 17633.58 \\
& & DRAM (\si{\joule}) & \underline{66.61} & 1.55 & 524.23 & 6.66 & 5472.08 & 131.74 & 44645.81 & 13256.05 \\
& & Time (\si{\second}) & \underline{15.68} & 0.10 & 105.04 & 0.41 & \textbf{1085.80} & 9.25 & 10414.56 & 653.39 \\
& & Memory (\si{\mega\byte}) & \underline{12.50} & 0.70 & 87.94 & 1.22 & 866.82 & 0.00 & 8667.12 & 0.00 \\
\cmidrule{2-11}
&\multirow{4}{*}{LHS} & CPU (\si{\joule}) & \underline{869.28} & 3.33 & 6262.98 & 45.77 & 61292.69 & 437.50 & 619914.82 & 20875.12 \\
& & DRAM (\si{\joule}) & \underline{66.36} & 0.85 & 523.87 & 9.93 & 5538.34 & 100.44 & 39263.81 & 22132.45 \\
& & Time (\si{\second}) & \underline{15.75} & 0.07 & 105.29 & 0.80 & 1095.78 & 7.12 & 10425.02 & 604.48 \\
& & Memory (\si{\mega\byte}) & \underline{12.62} & 0.54 & 89.54 & 2.11 & 866.82 & 0.00 & 8667.24 & 0.30 \\

\bottomrule
\end{tabular}
\begin{tablenotes}
\item[*] The significantly smaller values between the two sampling methods are highlighted in \textbf{bold}, while the minimum values among different dataset sizes are \underline{underlined}.
\end{tablenotes}
}
\end{threeparttable}
\end{table}

Table~\ref{tab:results_training} summarizes the computational cost of training the DNN surrogates under two sampling strategies and four dataset sizes. In this table, the Mann-Whitney U test is applied to identify statistically significant differences between the two sampling methods or different dataset sizes. Values that are significantly smaller between sampling methods are highlighted in bold, while the minimum values among different dataset sizes are underlined.

Overall, the computational cost increases as the dataset size grows. This trend is consistently observed across all instances and sampling strategies, reflecting the higher computational burden associated with training on larger datasets. 

When comparing the problem instances, the Stockholm and Paris instances require more computational resources than \Malaga. This behavior can be attributed to their larger problem dimensionality, which increases the number of trainable parameters and consequently, the training complexity of DNN.

Between the sampling strategies, no notable differences are observed across any dataset size or computational metric. CPU and DRAM energy, time, and memory usage remain nearly identical for both methods. This indicates that the choice of sampling strategy does not significantly affect training efficiency.

\begin{figure}[!tb]
    \centering
    \begin{minipage}[t]{0.32\linewidth}
        \includegraphics[width=\linewidth]{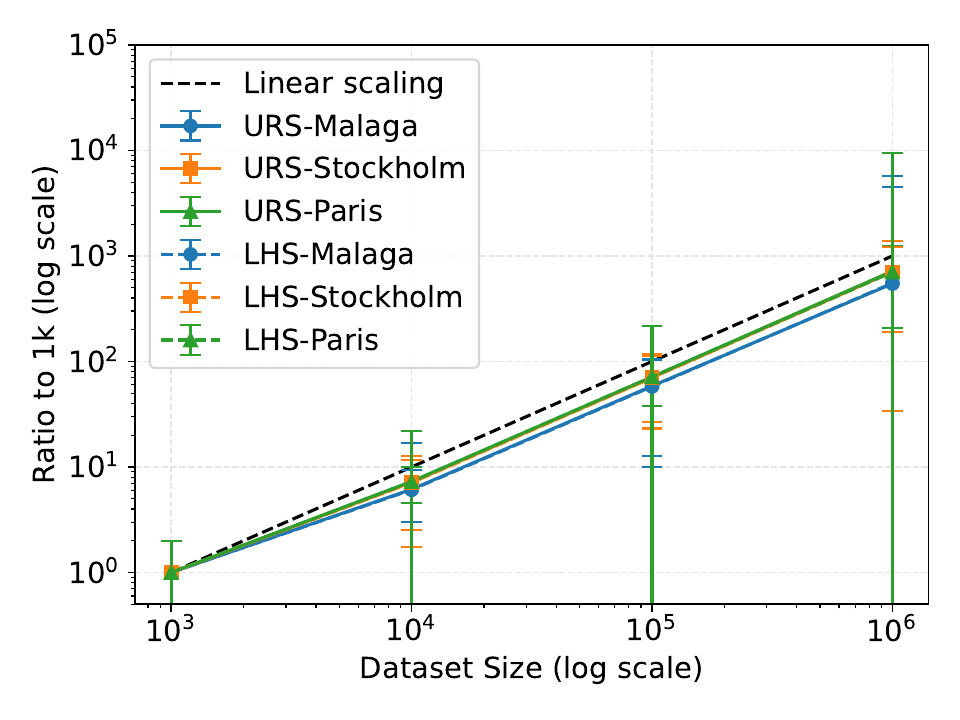}
        \subcaption{Energy (CPU+DRAM)}
    \end{minipage}
    \hfill
    \begin{minipage}[t]{0.32\linewidth}
        \includegraphics[width=\linewidth]{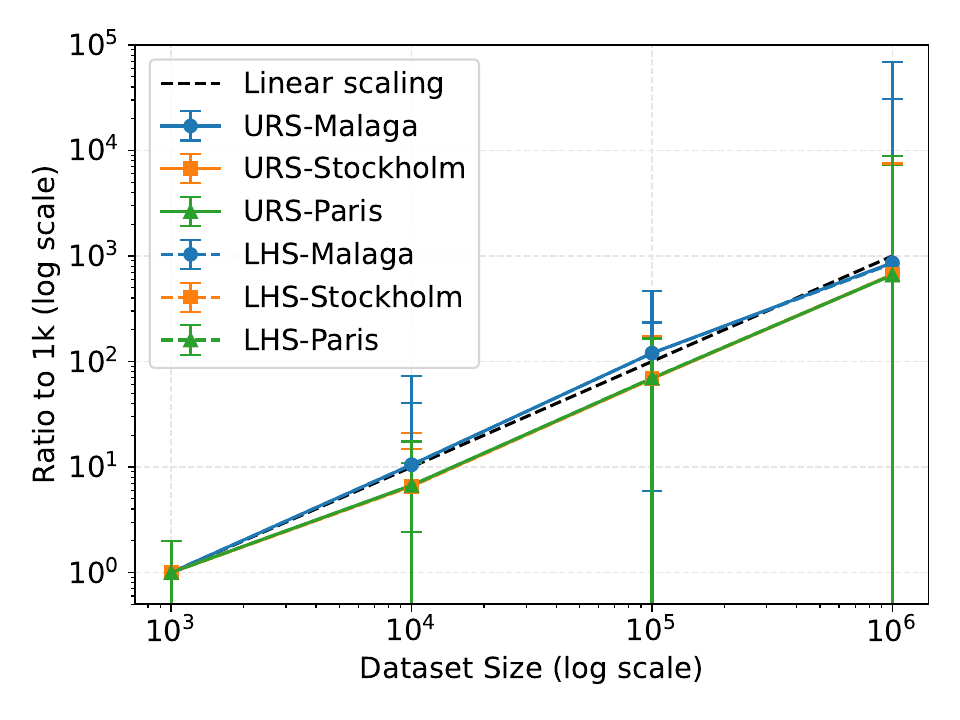}
        \subcaption{Time}
    \end{minipage}
    \hfill
    \begin{minipage}[t]{0.32\linewidth}
        \includegraphics[width=\linewidth]{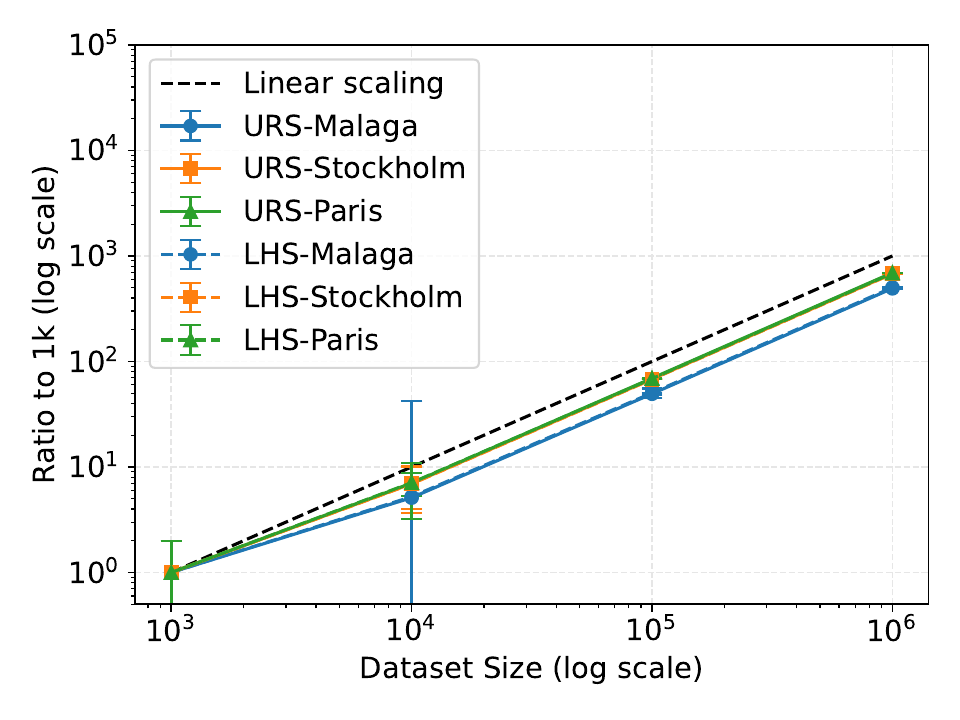}
        \subcaption{Memory}
    \end{minipage}
    \caption{Training metrics normalized to the 1k dataset}
    \label{fig:training_metrics_ratio}
\end{figure}
Fig.~\ref{fig:training_metrics_ratio} presents the training metrics normalized by the 1k dataset case. The horizontal axis represents the dataset size, while the vertical axis displays the normalized values of each metric on a logarithmic scale. The dashed black line represents the ideal linear scaling baseline. Note that the energy values correspond to the combined energy consumption of the CPU and DRAM.

The results confirm that the computational cost increases with dataset size; however, the growth trend is not uniform across metrics. Both energy and memory usage exhibit sublinear scaling. This means that although larger datasets require more resources, the increase is not proportional, reflecting improved training efficiency at larger scales. In contrast, execution time follows a pattern closer to linear growth, especially in the \Malaga instance. For Stockholm and Paris, time scaling is slightly sublinear, suggesting that the training process may benefit from greater efficiency at larger problems.

Considering the results in Table~\ref{tab:results_mean_var}, LHS yields marginally more uniform sampling and slightly lower variance, but these advantages vanish in large datasets where both methods achieve equivalent PTME performance.

\subsection{Inference PTME Analysis}
\subsubsection{Computational Efficiency}

The inference results demonstrate that computational costs remain relatively stable regardless of the training dataset sizes. Across all three problem instances and both sampling strategies, CPU energy consumption stays in the range of approximately 2.8--3.0 \si{\joule}, DRAM energy remains around 0.16--0.20 \si{\joule}, execution time is consistently close to 70 \si{\milli\second}, and memory usage remains within 116--119 \si{\kilo\byte}. 

This stability confirms that once the surrogate model is trained, inference can be performed with constant and low computational overhead, independent of the size of the training dataset and the sampling strategy. 

\subsubsection{Precision (error)}
\begin{table}[tb]
\centering
\caption{Predictive error (MAPE and RMSE) for different dataset sizes}
\label{tab:results_accuracy}
\begin{threeparttable}
\fontsize{6}{6}\selectfont{
\begin{tabular}{lllrr@{\quad}rr@{\quad}rr@{\quad}rr}
\toprule
\multirow{2}{*}{Instance}&\multirow{2}{*}{Method} & \multirow{2}{*}{Metric} & \multicolumn{2}{c}{1k} & \multicolumn{2}{c}{10k} & \multicolumn{2}{c}{100k} & \multicolumn{2}{c}{1M} \\
 &&  & Avg. & Stdv. & Avg. & Stdv. & Avg. & Stdv. & Avg. & Stdv. \\
 \midrule
\multirow{4}{*}{\Malaga}&\multirow{2}{*}{URS} & MAPE (\%, $\downarrow$) & 30.54 & 1.92 & 25.88 & 2.92 & 13.55 & 0.93 & \underline{9.37} & 0.25 \\
 && RMSE ($\downarrow$) & 0.56 & 0.06 & 0.38 & 0.01 & 0.25 & 0.01 & \underline{0.18} & 0.01 \\
\cmidrule{2-11}
&\multirow{2}{*}{LHS} & MAPE (\%, $\downarrow$) & 32.33 & 5.79 & 24.00 & 2.02 & 13.69 & 0.65 & \underline{9.32} & 0.35 \\
 && RMSE ($\downarrow$) & 0.55 & 0.05 & 0.39 & 0.02 & 0.26 & 0.01 & \underline{0.18} & 0.01 \\
  \midrule
\multirow{4}{*}{Stockholm} & \multirow{2}{*}{URS} & MAPE (\%, $\downarrow$) & 31.53 & 2.13 & 28.79 & 3.01 & 18.12 & 0.73 & \underline{14.21} & 0.54 \\
 && RMSE ($\downarrow$) & 0.51 & 0.02 & 0.43 & 0.03 & 0.30 & 0.01 & \underline{0.23} & 0.01 \\
\cmidrule{2-11}
&\multirow{2}{*}{LHS} & MAPE (\%, $\downarrow$) & 30.97 & 1.95 & \textbf{25.43} & 0.96 & \textbf{16.30} & 0.53 & \underline{14.05} & 1.04 \\
& & RMSE ($\downarrow$) & 0.51 & 0.02 & 0.43 & 0.03 & \textbf{0.28} & 0.01 & \underline{0.23} & 0.01 \\
 \midrule
\multirow{4}{*}{Paris}&\multirow{2}{*}{URS} & MAPE (\%, $\downarrow$) & 33.01 & 4.21 & 29.04 & 2.25 & 17.78 & 0.85 & \underline{13.79} & 0.93 \\
& & RMSE ($\downarrow$) & 0.51 & 0.02 & 0.43 & 0.01 & 0.30 & 0.01 & \underline{0.23} & 0.01 \\
\cmidrule{2-11}
&\multirow{2}{*}{LHS} & MAPE (\%, $\downarrow$) & 32.71 & 3.06 & \textbf{26.37} & 0.89 & 17.15 & 0.89 & \underline{13.64} & 0.42 \\
& & RMSE ($\downarrow$) & 0.52 & 0.03 & 0.44 & 0.02 & 0.30 & 0.01 & \underline{0.24} & 0.01 \\
\bottomrule
\end{tabular}
\begin{tablenotes}
\item[*] The significantly small values between the two sampling methods are highlighted in \textbf{bold}, while the minimum values among different dataset sizes are \underline{underlined}.
\end{tablenotes}
}
\end{threeparttable}
\end{table}

Table~\ref{tab:results_accuracy} presents the predictive error of the trained surrogates. Similar to Table~\ref{tab:results_training}, significantly better results are in bold or underlined.

Across all problem instances, the predictive error decreases monotonically as the dataset size increases. This trend holds for both MAPE and RMSE, demonstrating that larger training datasets yield more accurate surrogate models.

Among the instances, the \Malaga instance achieves the lowest error when trained with 100k or more samples. This behavior aligns with its relatively lower dimensionality, which makes the underlying function easier to approximate and reduces the sample complexity of surrogate learning. In contrast, the Stockholm and Paris instances exhibit higher residual error, reflecting the greater difficulty of modeling higher-dimensional search spaces.

Regarding sampling strategies, a notable difference emerges at the 10k dataset size, where LHS outperforms URS across all instances and metrics. This indicates that LHS provides better coverage of the input domain when the dataset is still modest in size, thereby improving learning accuracy. However, for the 1k and 100k--1M cases, the performance gap narrows considerably, suggesting that sufficient data eventually compensates for differences in sampling strategy, enabling both approaches to achieve comparable accuracy.

\begin{figure}[tb]
\begin{minipage}[c]{0.65\linewidth}
    \centering
    \begin{minipage}[t]{0.48\linewidth}
        \includegraphics[width=\linewidth]{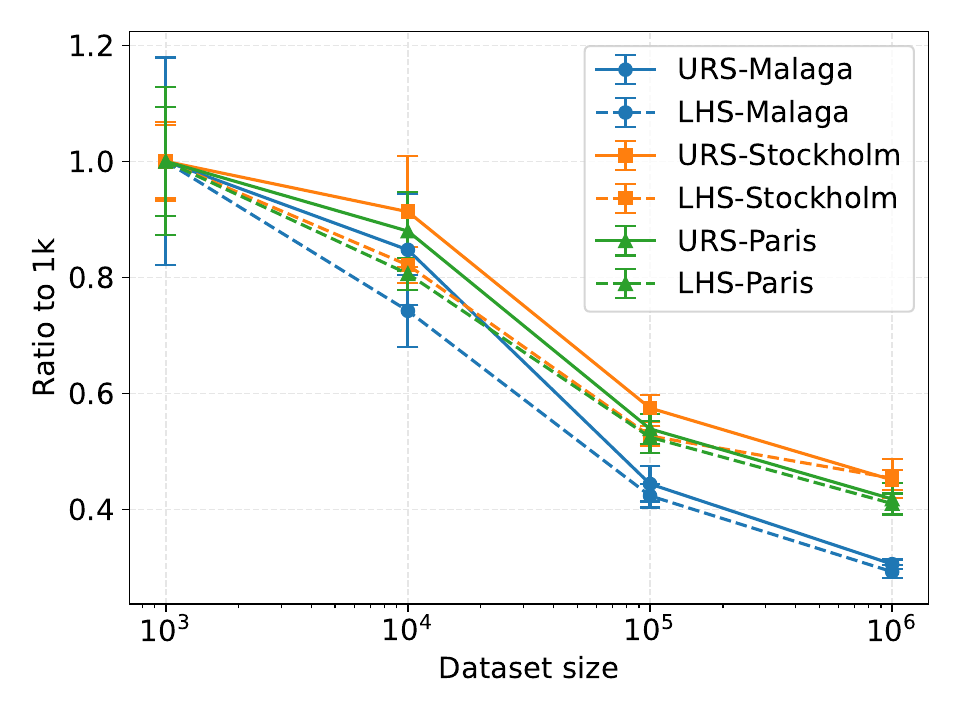}
        \subcaption{MAPE}
    \end{minipage}
    \begin{minipage}[t]{0.48\linewidth}
        \includegraphics[width=\linewidth]{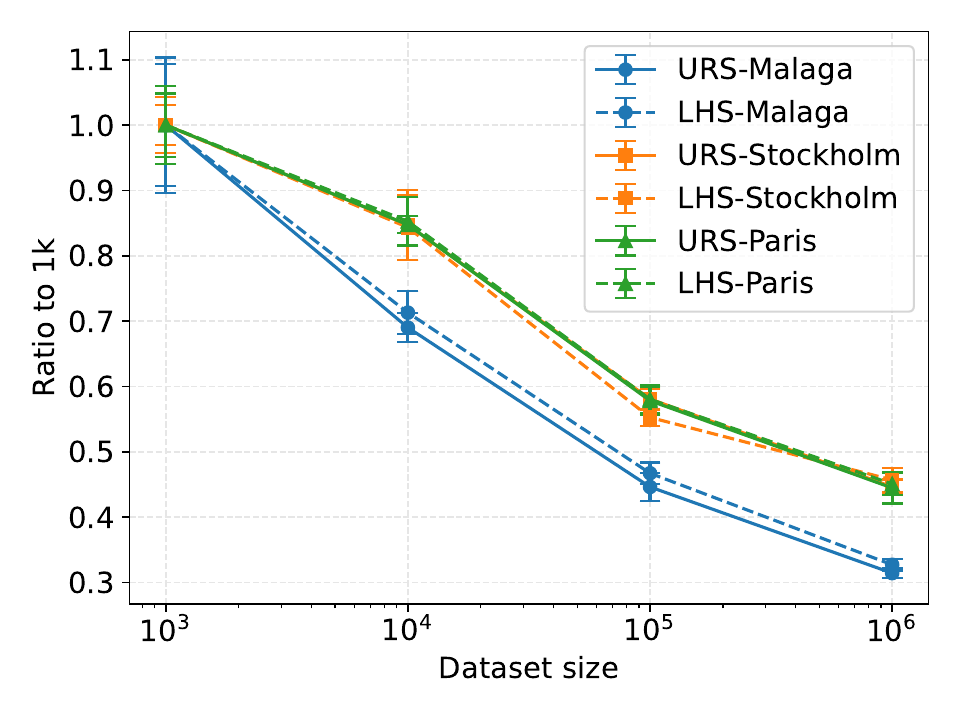}
        \subcaption{RMSE}
    \end{minipage}
    \caption{Error metrics normalized to 1k dataset}
    \label{fig:inference_accuracy_ratio}
    \end{minipage}
    \hfill
    \begin{minipage}[c]{0.32\linewidth}
    \centering
    \includegraphics[width=\linewidth]{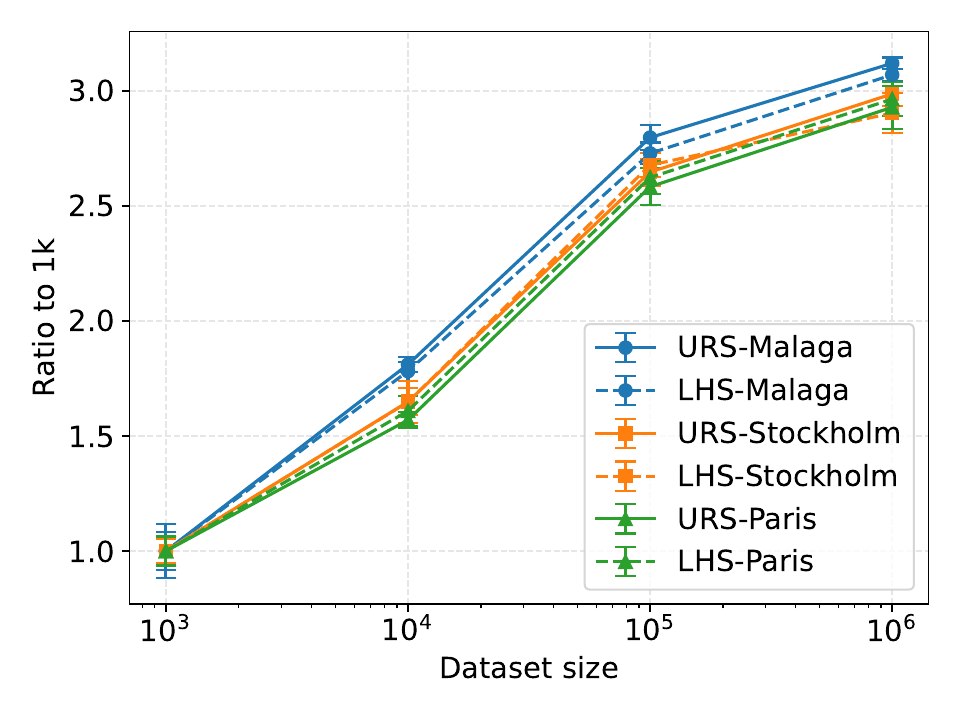}
    \caption{Kendall's $\tau_a$ normalized to 1k dataset}
    \label{fig:inference_rank_correlation_ratio}
    \end{minipage}
\end{figure}
Fig.~\ref{fig:inference_accuracy_ratio} illustrates the normalized error metrics with respect to the dataset size of 1k. The horizontal axis represents the dataset size on a logarithmic scale, while the vertical axis shows the normalized values of each metric.

The results clearly show that increasing the dataset size reduces learning error, with a 60–70\% improvement from 1k to 1M samples. However, the gains are nonlinear: the improvement from 1k to 10k is modest, the largest reduction occurs between 10k and 100k, and gains beyond 100k become more gradual.

This is because small datasets only allow learning of coarse patterns, and meaningful details emerge only once the data becomes dense enough, around 100k samples. Beyond this point, most of the structure is already captured, so additional samples yield diminishing returns, making the improvement from 100k to 1M smaller than the jump from 10k to 100k.

\subsubsection{Precision (rank correlation)}
We evaluate the rank correlation using Kendall’s $\tau_a$ and $\tau_b$. Kendall’s $\tau_a$ and $\tau_b$ exhibit nearly identical values across all settings, because the objective function used in this study is continuous and rarely produces tied rankings. Therefore, we report and discuss them jointly.

Rank correlation improves consistently as the dataset size increases. For the \Malaga instance, $\tau$ rises from approximately 0.26 at 1k samples to around 0.47 at 10k, 0.72--0.73 at 100k, and approximately 0.80--0.81 at 1M. Stockholm exhibits a similar trend, improving from approximately 0.24 at 1k to 0.39 at 10k, 0.63--0.65 at 100k, and around 0.70–0.71 at 1M. Paris follows the same pattern, increasing from about 0.25 to 0.39, 0.63--0.64, and finally 0.71--0.72 over the same sample sizes. These results demonstrate that larger datasets substantially enhance the surrogate’s ability to preserve solution rankings.

The \Malaga instance achieves slightly higher correlation values than Stockholm and Paris when using datasets larger than 10k, which can be attributed to its lower dimensionality and comparatively simpler learning landscape.

When comparing the two sampling strategies, consistent with earlier findings, no meaningful differences are observed between URS and LHS at any dataset size, indicating that both strategies are similarly effective for rank-based surrogate learning once a sufficiently large dataset is used.

Fig.~\ref{fig:inference_rank_correlation_ratio} illustrates the normalized rank correlation metric (Kendall's $\tau_a$) with respect to the dataset size of 1k. The horizontal axis represents dataset size on a logarithmic scale, while the vertical axis shows the normalized metric values. Note that because Kendall's $\tau_a$ and $\tau_b$ are identical, we only present Kendall's~$\tau_a$.

The results clearly indicate that increasing the dataset size improves rank correlation, with approximately a threefold increase observed from 1k to 1M samples. However, similar to the error metrics, the improvement is nonlinear across data scales. The gain from 1k to 10k is modest, whereas the most pronounced improvement occurs between 10k and 100k. Beyond 100k, the correlation continues to increase but with diminishing returns.

\section{Surrogate-Assisted Optimization Results}\label{sec:search}
The previous PTME results guide the below integration of surrogates into SAPSO: the observed constant inference cost enables scalable optimization, while the sublinear training scaling supports periodic model retraining with manageable overhead.

\subsection{Surrogate-Assisted Particle Swarm Optimization Setup}
\begin{algorithm}[tb]
\caption{A pseudo-code of DNN surrogate-assisted PSO (SAPSO)}
\label{alg:sapso}
\DontPrintSemicolon
Load pretrained DNN surrogate model $\hat{f}$\;
Randomly generate $N$ solutions as the initial swarm $P_0$\;
Evaluate all solutions using the surrogate model $\hat{f}$\;
Set the global and personal best particles\;
$FE \gets N$, $g \gets 0$\;
\While{$FE < MaxFE$}{
    \ForEach{particle $\bm{x}_g^i$}{
        Update velocity $\bm{v}_{g+1}^i\leftarrow w\bm{v}_g^i + \phi_1U(0, 1)(\bm{p}_g^i-\bm{x}_g^i)+\phi_2 U(0, 1)(\bm{b}_g-\bm{x}_g^i)$\;
        Update position $\bm{x}_{g+1}^i\leftarrow \bm{x}_{g}^i + \bm{v}_{g+1}^i$\;
        Predict objective function value of $\bm{x}_{g+1}^i$ using $\hat{f}$\;
        Update the personal best particle $\bm{p}_{g+1}^i$\;
    }
    Update the global best particle $\bm{b}_{g+1}$\;
    $FE \gets FE + N, g \gets g + 1$\;
}
Evaluate the best solution in $P_g$ using actual expensive function $f$\;
\end{algorithm}
In this experiment, we integrate the pretrained surrogate models into a particle swarm optimization (PSO) algorithm~\cite{Kennedy1995}, resulting in surrogate-assisted PSO (SAPSO) as outlined in Algorithm~\ref{alg:sapso}. The key modification involves using the pretrained surrogate model to predict the objective values of particles, thereby replacing the expensive function evaluations. Because we obtained ten different surrogate models in the previous experiments, we ran the SAPSO algorithm ten times for each combination of sampling strategy and dataset size.

In this experiment, we use the parameter settings used in the previous studies~\cite{Harada2024Energy,Segredo2019}. In particular, we set the local and global coefficients ($\phi_1$ and $\phi_2$) to 2.05, the velocity truncation factor ($\lambda$) to 0.5, and the inertia weight to decrease linearly from 0.5 to 0.1 over the optimization. The population (swarm) size is 100, and a maximum of 30,000 surrogate evaluations is allowed.

As a baseline for comparison, we also implement PSO without surrogates, using the same parameters as in SAPSO, while executing 30,000 actual objective function evaluations.

\subsection{Optimization Results}
\begin{figure}[tb]
\centering
\begin{minipage}[t]{0.32\linewidth}
\includegraphics[width=\linewidth]{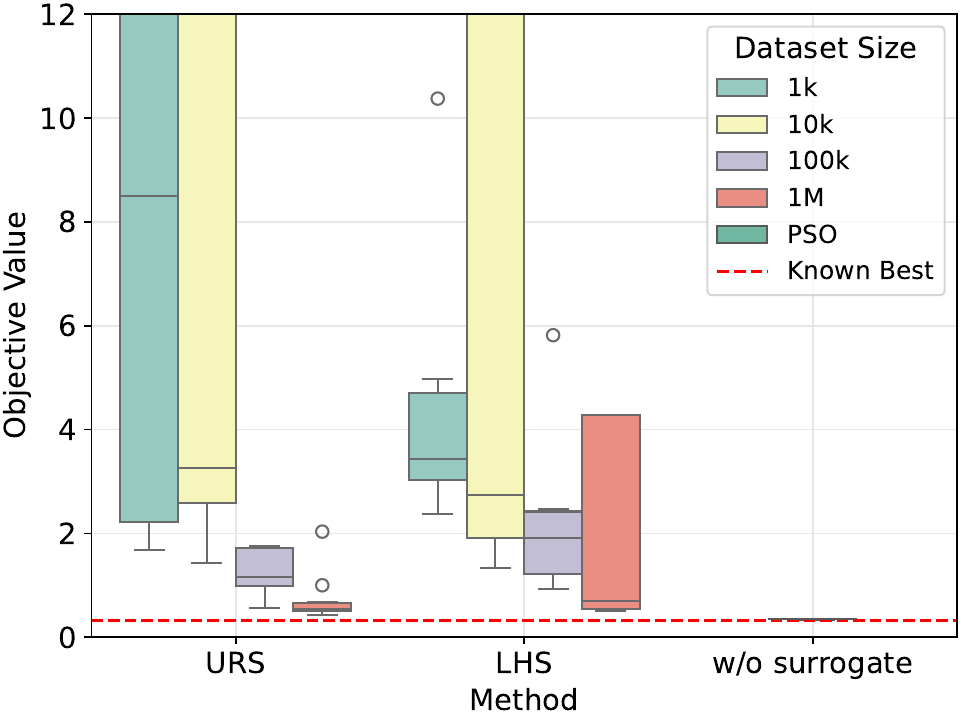}
\subcaption{\Malaga}
\end{minipage}
\hfill
\begin{minipage}[t]{0.32\linewidth}
\includegraphics[width=\linewidth]{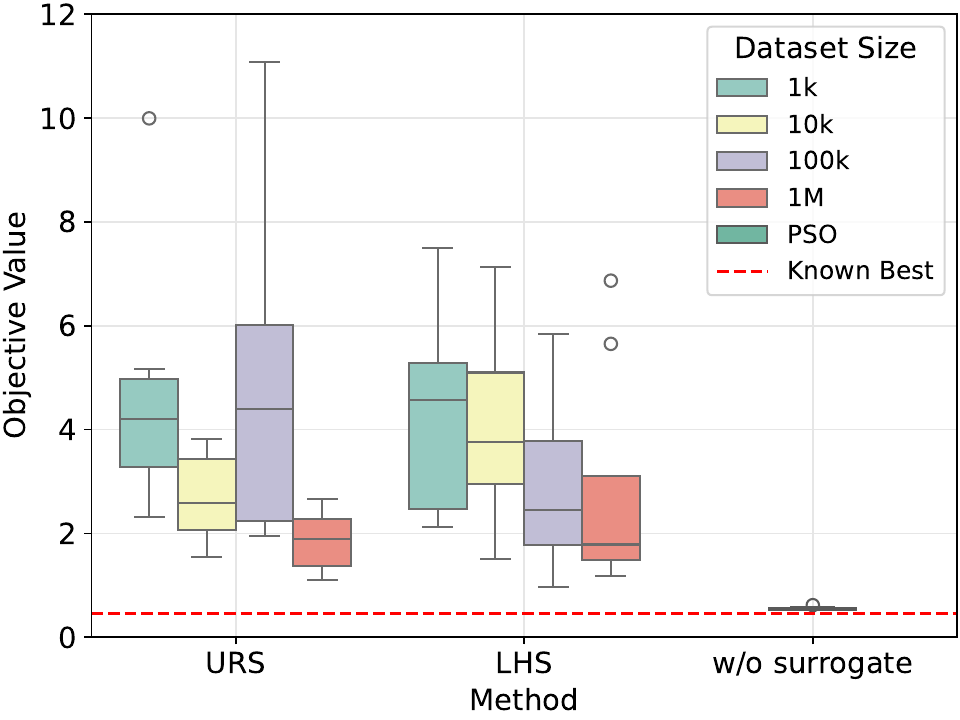}
\subcaption{Stockholm}
\end{minipage}
\hfill
\begin{minipage}[t]{0.32\linewidth}
\includegraphics[width=\linewidth]{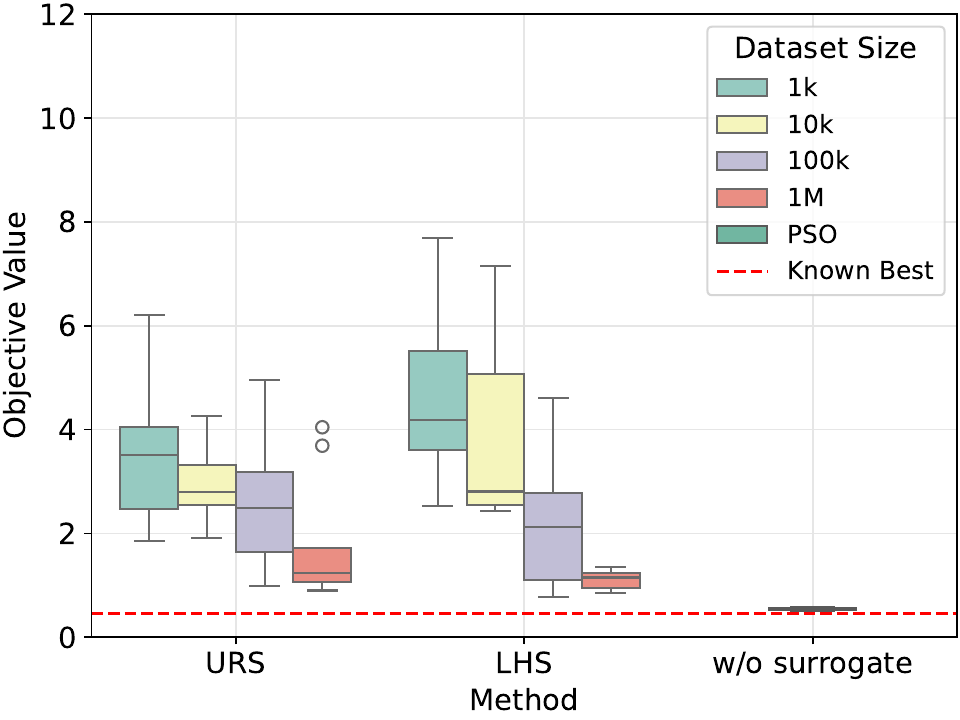}
\subcaption{Paris}
\end{minipage}
\caption{Best objective value found by SAPSO and PSO without surrogates}
\label{fig:best_objective_value}
\end{figure}
Fig.~\ref{fig:best_objective_value} illustrates the boxplot of the best objective values found by SAPSO using pretrained surrogates. 
The horizontal axis indicates the different sampling strategies and PSO without surrogates, while the vertical axis represents the best objective value found. The horizontal dashed line indicates the known best objective value for each problem instance. Although some results include outliers with values in the tens or even hundreds, the vertical axis is limited for clarity of visualization.

From the results, we observe that as the dataset size increases, SAPSO performance improves across all three problem instances and both sampling strategies. This indicates that larger training datasets yield more accurate surrogate models, which, in turn, improve SAPSO's optimization performance.

While SAPSO achieves slightly lower precision than full evaluations in PSO, it substantially reduces computational cost and energy, showing potential when combined with, e.g., adaptive retraining or hybrid strategies.

\section{Conclusions and Future Works}\label{sec:conclusion}

This study applied the PTME methodology to analyze deep learning surrogates in terms of Precision, Time, Memory, and Energy, providing a quantitative and reproducible framework for surrogate evaluation in complex optimization.

Experiments confirmed that larger dataset sizes yield higher precision and better Kendall’s rank correlation, making surrogates more consistent with the original problem fitness. This property is essential for their use inside solvers that rely on pairwise comparisons to guide the search toward optimal solutions. Although both $\tau_a$ and $\tau_a$ were computed, no significant differences were found due to the continuous fitness function of the studied instances.

A sublinear relationship was observed between dataset size and training time, memory, and energy consumption, indicating that the computational cost increases slower than linearly while accuracy improves. This scaling behavior supports the feasibility of training accurate surrogates with acceptable resource use.

Inference remained constant and low-cost across all configurations—both in time, memory, and energy—making it the most efficient stage for integrating surrogates within optimization algorithms. Therefore, minimizing re-training during search is recommended to maintain computational efficiency.

No significant differences were detected between uniform random sampling (URS) and Latin hypercube sampling (LHS) in any PTME dimension. LHS achieved marginally higher uniformity and accuracy in small and medium datasets, but both approaches converged to similar results at larger scales.

The surrogate-assisted PSO experiments demonstrated surrogates improve the original performance trends across different city instances, at the price of a slightly lower precision.

Overall, the PTME methodology proved effective for balancing numerical precision and computational cost in surrogate evaluation. It offers a solid baseline for assessing new models in real-world optimization. Future work will extend PTME to adaptive and multi-fidelity surrogates, incorporating online re-training and energy-aware optimization mechanisms in the path for a greener AI.

\begin{credits}

\subsubsection{\discintname}
The authors declare that they have no known competing interests or personal relationships that could have appeared to influence the work reported in this paper. This work does not represent the interest or opinion of any of the author's affiliated institutions.
\end{credits}

%
%
%
\bibliographystyle{splncs04}
\bibliography{references}

@Article{Jamil2022,
author={Jamil, Mohammad Newaj
and Kor, Ah-Lian},
title={Analyzing energy consumption of nature-inspired optimization algorithms},
journal={Green Technology, Resilience, and Sustainability},
year={2022},
month={Jan},
day={27},
volume={2},
number={1},
pages={1},
issn={2731-3425},
doi={10.1007/s44173-021-00001-9},
}

@article{Diaz2022,
author = {Josefa Díaz-\'{A}lvarez and Pedro A Castillo and Francisco Fern\'{a}ndez de Vega and Francisco Ch\'{a}vez and Jorge Alvarado},
title ={Population size influence on the energy consumption of genetic programming},

journal = {Meas. and Control},
volume = {55},
number = {1-2},
pages = {102-115},
year = {2022},
doi = {10.1177/00202940211064471}
}

@Article{Abdelhafez2019,
author={Abdelhafez, Amr
and Alba, Enrique
and Luque, Gabriel},
title={A component-based study of energy consumption for sequential and parallel genetic algorithms},
journal={The Journal of Supercomputing},
year={2019},
month={Oct},
day={01},
volume={75},
number={10},
pages={6194-6219},
issn={1573-0484},
doi={10.1007/s11227-019-02843-4},
}

@article{Ferro2023,
   author = {Mariza Ferro and Gabrieli D. Silva and Felipe B. de Paula and Vitor Vieira and Bruno Schulze},
   doi = {10.1002/cpe.6815},
   issn = {15320634},
   issue = {17},
   journal = {Concurrency and Comp.: Pract. and Exp.},
   month = {8},
   publisher = {John Wiley and Sons Ltd},
   title = {Towards a sustainable {AI}: A case study of energy efficiency in decision tree algorithms},
   volume = {35},
   year = {2023},
}

@INPROCEEDINGS{David2010,
  author={David, Howard and Gorbatov, Eugene and Hanebutte, Ulf R. and Khanna, Rahul and Le, Christian},
  booktitle={2010 International Symposium on Low-Power Electronics and Design}, 
  title={{RAPL}: Memory power estimation and capping}, 
  year={2010},
  volume={},
  number={},
  pages={189-194},
  doi={10.1145/1840845.1840883}}

@misc{anthony2020carbontracker,
  title={Carbontracker: Tracking and Predicting the Carbon Footprint of Training Deep Learning Models},
  author={Lasse F. Wolff Anthony and Benjamin Kanding and Raghavendra Selvan},
  howpublished={ICML Workshop on Challenges in Deploying and monitoring Machine Learning Systems},
  month={July},
  note={arXiv:2007.03051},
  year={2020}}

@article{GARCIANIETO2012274,
title = {Swarm intelligence for traffic light scheduling: Application to real urban areas},
journal = {Engineering Applications of Artificial Intelligence},
volume = {25},
number = {2},
pages = {274-283},
year = {2012},
issn = {0952-1976},
doi = {10.1016/j.engappai.2011.04.011},
author = {J. García-Nieto and E. Alba and A. {Carolina Olivera}},
}

@ARTICLE{Paul2023Green,
  author={Paul, Showmick Guha and Saha, Arpa and Arefin, Mohammad Shamsul and Bhuiyan, Touhid and Biswas, Al Amin and Reza, Ahmed Wasif and Alotaibi, Naif M. and Alyami, Salem A. and Moni, Mohammad Ali},
  journal={IEEE Access}, 
  title={A Comprehensive Review of Green Computing: Past, Present, and Future Research}, 
  year={2023},
  volume={11},
  number={},
  pages={87445-87494},
  doi={10.1109/ACCESS.2023.3304332}}

@Article{khalfi2023metaheuristics,
author={Khalfi, Souheila
and Caraffini, Fabio
and Iacca, Giovanni},
title={Metaheuristics in the Balance: A Survey on Memory-Saving Approaches for Platforms with Seriously Limited Resources},
journal={International Journal of Intelligent Systems},
year={2023},
month={Nov},
day={04},
publisher={Hindawi},
volume={2023},
pages={5708085},
issn={0884-8173},
doi={10.1155/2023/5708085},
}

@ARTICLE{Segredo2019,
  author={Segredo, Eduardo and Luque, Gabriel and Segura, Carlos and Alba, Enrique},
  journal={IEEE Access}, 
  title={Optimising Real-World Traffic Cycle Programs by Using Evolutionary Computation}, 
  year={2019},
  volume={7},
  number={},
  pages={43915-43932},
  doi={10.1109/ACCESS.2019.2908562}}

@article{VILLAGRA2020101085,
title = {A better understanding on traffic light scheduling: New cellular GAs and new in-depth analysis of solutions},
journal = {Journal of Computational Science},
volume = {41},
pages = {101085},
year = {2020},
issn = {1877-7503},
doi = {https://doi.org/10.1016/j.jocs.2020.101085},
url = {https://www.sciencedirect.com/science/article/pii/S1877750319302169},
author = {Andrea Villagra and Enrique Alba and Gabriel Luque}
}

@misc{Fang2005,
  title = {Design and Modeling for Computer Experiments},
  ISBN = {9780429143762},
  url = {http://dx.doi.org/10.1201/9781420034899},
  DOI = {10.1201/9781420034899},
  publisher = {Chapman and Hall/CRC},
  author = {Fang,  Kai-Tai and Li,  Runze and Sudjianto,  Agus},
  year = {2005},
  month = oct 
}

@INPROCEEDINGS{Harada2024Energy,
  author={Harada, Tomohiro and Alba, Enrique and Luque, Gabriel},
  booktitle={2024 IEEE Congress on Evolutionary Computation (CEC)}, 
  title={Energy and Quality of Surrogate-Assisted Search Algorithms: a First Analysis}, 
  year={2024},
  volume={},
  number={},
  pages={1-8},
  doi={10.1109/CEC60901.2024.10611758}}

@INPROCEEDINGS{Kennedy1995,
  author={Kennedy, J. and Eberhart, R.},
  booktitle={Proceedings of ICNN'95 - International Conference on Neural Networks}, 
  title={Particle swarm optimization}, 
  year={1995},
  volume={4},
  number={},
  pages={1942-1948 vol.4},
  doi={10.1109/ICNN.1995.488968}}

@inproceedings{SUMO2018,
          title = {Microscopic Traffic Simulation using SUMO},
         author = {Pablo Alvarez Lopez and Michael Behrisch and Laura Bieker-Walz and Jakob Erdmann and Yun-Pang Fl{\"o}tter{\"o}d and Robert Hilbrich and Leonhard L{\"u}cken and Johannes Rummel and Peter Wagner and Evamarie Wie{\ss}ner},
      publisher = {IEEE},
      booktitle = {The 21st IEEE International Conference on Intelligent Transportation Systems},
           year = {2018},
        journal = {IEEE Intelligent Transportation Systems Conference (ITSC)},
            url = {https://elib.dlr.de/124092/}
 }

@Article{Balderas2021,
author={Balderas, David
and Ortiz, Alexandro
and M{\'e}ndez, Efra{\'i}n
and Ponce, Pedro
and Molina, Arturo},
title={Empowering Digital Twin for Industry 4.0 using metaheuristic optimization algorithms: case study PCB drilling optimization},
journal={The International Journal of Advanced Manufacturing Technology},
year={2021},
month={Mar},
day={01},
volume={113},
number={5},
pages={1295-1306},
issn={1433-3015},
doi={10.1007/s00170-021-06649-8},
url={https://doi.org/10.1007/s00170-021-06649-8}
}

@incollection{DAVIS2018979,
title = {Efficient Surrogate Model Development: Impact of Sample Size and Underlying Model Dimensions},
editor = {Mario R. Eden and Marianthi G. Ierapetritou and Gavin P. Towler},
series = {Computer Aided Chemical Engineering},
publisher = {Elsevier},
volume = {44},
pages = {979-984},
year = {2018},
booktitle = {13th International Symposium on Process Systems Engineering (PSE 2018)},
issn = {1570-7946},
doi = {https://doi.org/10.1016/B978-0-444-64241-7.50158-0},
url = {https://www.sciencedirect.com/science/article/pii/B9780444642417501580},
author = {Sarah E. Davis and Selen Cremaschi and Mario R. Eden}
}

@article{WASEEM2025344,
title = {Machine learning-enhanced digital twins for predictive analytics in battery pack assembly},
journal = {Journal of Manufacturing Systems},
volume = {80},
pages = {344-355},
year = {2025},
issn = {0278-6125},
doi = {https://doi.org/10.1016/j.jmsy.2025.03.007},
url = {https://www.sciencedirect.com/science/article/pii/S0278612525000688},
author = {Muhammad Waseem and Changbai Tan and Seog-Chan Oh and Jorge Arinez and Qing Chang}
}

@article{HWANG201874,
title = {A fast-prediction surrogate model for large datasets},
journal = {Aerospace Science and Technology},
volume = {75},
pages = {74-87},
year = {2018},
issn = {1270-9638},
doi = {https://doi.org/10.1016/j.ast.2017.12.030},
url = {https://www.sciencedirect.com/science/article/pii/S1270963817309240},
author = {John T. Hwang and Joaquim R.R.A. Martins}
}

@article{cheng2024review,
  title={A review of data-driven surrogate models for design optimization of electric motors},
  author={Cheng, Mengyu and Zhao, Xing and Dhimish, Mahmoud and Qiu, Wangde and Niu, Shuangxia},
  journal={IEEE Transactions on Transportation Electrification},
  volume={10},
  number={4},
  pages={8413--8431},
  year={2024},
  publisher={IEEE}
}

@Article{Harada2023,
author={Harada, Tomohiro},
title={A pairwise ranking estimation model for surrogate-assisted evolutionary algorithms},
journal={Complex {\&} Intelligent Systems},
year={2023},
month={Dec},
day={01},
volume={9},
number={6},
pages={6875-6890},
issn={2198-6053},
doi={10.1007/s40747-023-01113-4},
url={https://doi.org/10.1007/s40747-023-01113-4}
}

@article{BLIEK2023110744,
title = {Benchmarking surrogate-based optimisation algorithms on expensive black-box functions},
journal = {Applied Soft Computing},
volume = {147},
pages = {110744},
year = {2023},
issn = {1568-4946},
doi = {https://doi.org/10.1016/j.asoc.2023.110744},
url = {https://www.sciencedirect.com/science/article/pii/S1568494623007627},
author = {Laurens Bliek and Arthur Guijt and Rickard Karlsson and Sicco Verwer and Mathijs {de Weerdt}}
}
\end{document}